\documentclass{article}
\usepackage{ijcai25}

\usepackage[switch]{lineno}

\usepackage[hidelinks]{hyperref}
\usepackage{url}
\usepackage{times}
\usepackage{latexsym}
\usepackage[T1]{fontenc}
\usepackage[utf8]{inputenc}
\usepackage{microtype}
\usepackage{inconsolata}
\usepackage{graphicx}
\usepackage{multirow}
\usepackage{diagbox}
\usepackage{booktabs}
\usepackage{amsmath, amssymb, bm}
\usepackage{algorithm}
\usepackage{algorithmic}
\floatname{algorithm}{Algorithm}
\usepackage{color}
\newcommand{\red}[1]{\textcolor{red}{#1}}

\newcommand{\blue}[1]
{\textcolor{blue}{#1}}
\usepackage{caption}
\title{Beyond Fixed Length: Bucket Pre-training is All You Need}

\author{
    Qing Yang\textsuperscript{\rm 1} \and
    Qiyao Peng\thanks{Corresponding author}\textsuperscript{\rm 2} \and 
    Hongtao Liu\textsuperscript{\rm 2} \and 
    Kai Liu\textsuperscript{\rm 3} \and 
    Bing Qin\textsuperscript{\rm 1} \And 
    Ting Liu\textsuperscript{\rm 1} \\
    \affiliations
    \textsuperscript{\rm 1}Harbin Institute of Technology, Harbin, Heilongjiang, China\\
    \textsuperscript{\rm 2}Du Xiaoman Financial Technology, Beijing, China\\
    \textsuperscript{\rm 3}Tianjin University, Tianjin, China\\
        \emails
    yangqing@duxiaoman.com, qypeng@tju.edu.cn, liuhongtao01@duxiaoman.com,
    kedixa@tju.edu.cn,
    {qinb,tliu}@ir.hit.edu.cn
}

\begin{document}

\maketitle

\begin{abstract}

Large Language Models (LLMs) have demonstrated exceptional performance across various tasks, with pre-training stage serving as the cornerstone of their capabilities.
However, the conventional fixed-length data composition strategy for pre-training presents several practical challenges. 
When using shorter sequences, documents are often truncated, potentially leading to information loss and affecting the model's ability to capture long-range dependencies. 
Conversely, longer sequences require concatenation of multiple documents, which can introduce noise and affect the natural document boundaries and semantic coherence as well as require substantial computational overhead.
To address these challenges, we first establish three quantitative metrics for evaluating data composition quality: padding ratio, truncation ratio, and concatenation ratio. 
Building upon these metrics, we propose a novel multi-bucket data composition method that transcends the fixed-length paradigm. 
Our approach adaptively organizes training data to achieve optimal composition quality as measured by the proposed metrics, offering a more flexible and efficient approach for pre-training.
We conduct extensive experiments and the results demonstrate that our proposed method significantly enhances both the efficiency and effectiveness of LLM pre-training.
Our proposed method has been adopted in the Du Xiaoman–XuanYuan series of financial large language models at \url{https://github.com/Duxiaoman-DI/XuanYuan}.

\end{abstract}
\section{Introduction}

Large Language Models (LLMs) have demonstrated unprecedented capabilities across diverse domains, including natural language processing, coding, and mathematical reasoning~\cite{brown2020language,ouyang2022training}.
The extraordinary performance of these models stems from their extensive pre-training on vast corpora of unlabeled text data. 
Given the immense scale of pre-training datasets, efficient and effective organization of input data during training becomes important. 
While ideally, each document would be processed as an independent training instance to preserve semantic integrity and eliminate noise, practical constraints such as computational efficiency, maximum context length, and batch processing requirements make this approach unfeasible. 

In current LLM pre-training, the dominant approach for organizing massive document collections is fixed-length method, which randomly concatenates documents and segments them into fixed-length sequences, typically 4096 or 8192 tokens, with special tokens marking document boundaries~\cite{rae2021scaling,zhang2022opt,touvron2023llama,pouransari2024dataset,yang2024qwen2,deepseekai2024deepseekv3technicalreport}. 
While this method has been widely adopted for its computational efficiency and batch processing compatibility, the arbitrary segmentation at fixed intervals introduces potential drawbacks. 
It often disrupts natural document boundaries and semantic coherence, potentially compromising the model's ability to learn long-range dependencies.

\begin{figure*}
\centering
\includegraphics[width=0.8\textwidth]{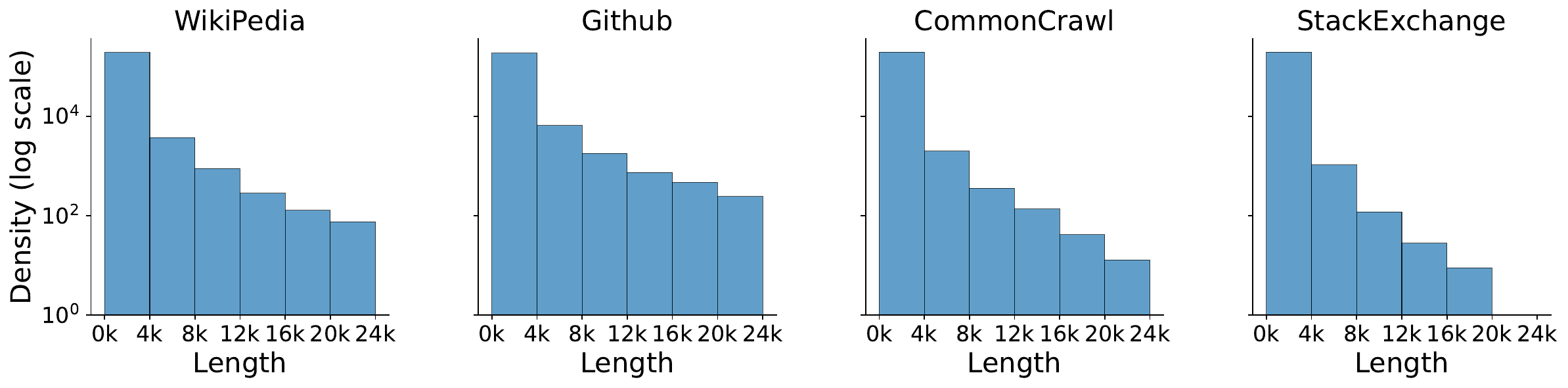}
\caption{Data Length Distribution.}
\label{fig:datadistribution} 
\end{figure*}

We analyze the length distributions across four representative pre-training corpora: WikiPedia, Github, CommonCrawl, StackExchange. 
Figure~\ref{fig:datadistribution} reveals that the majority of documents in these datasets are relatively short, predominantly within the 2k token range. 
This heterogeneous distribution exposes fundamental limitations of fixed-length training approaches. 
Using shorter sequence lengths (e.g., 2k) severely restricts the model's ability to capture long-range dependencies and understand broader context, which is crucial for complex reasoning and document-level comprehension. 
Furthermore, longer documents must be truncated into fixed-length segments, breaking the coherent flow of information and semantic relationships.
Conversely, longer lengths (e.g., 8k) not only necessitate extensive document concatenation, introducing artificial boundaries and potential noise, but also significantly increase computational overhead and slow down training speed. 
These issues collectively highlight the inherent inflexibility of fixed-length approaches in handling the diverse nature of real-world text data.

Recent works have attempted to address the challenges arising from document concatenation and splitting. 
Ding et al.~\cite{ding2024fewer} proposed Best-fit Packing, a length-aware combinatorial optimization approach to minimize unnecessary truncation. 
Similarly, Hadi et al.~\cite{pouransari2024dataset} developed a length-based bucketing method with curriculum sampling, utilizing binomial decomposition to create flexible training sequences.

While these methods show promising results, they lack a systematic framework for evaluating data composition quality in LLM pre-training. 
To bridge this gap, we propose three quantitative metrics:
\begin{itemize}
    \item \textit{Padding ratio}: The proportion of padding tokens in training sequences.
    \item \textit{Truncation ratio}: The percentage of truncated documents.
    \item \textit{Concatenation ratio}: The average number of documents combined in each training sample.
\end{itemize}
These metrics provide a comprehensive assessment of data composition quality, with lower values indicating better composition for pre-training.
A detailed discussion of these metrics is provided in Section~\ref{preliminary}.

Based on these observations and challenges, we propose to transcend the fixed-length training paradigm with a novel multi-bucket data composition method for LLM pre-training. 
Our approach adaptively organizes training data into length-specific buckets, significantly optimizing the aforementioned metrics to achieve higher quality data composition while maintaining training efficiency.
The main contributions of this work are:
\begin{itemize}
    \item \textbf{Data Composition Quality Metrics}: We design a comprehensive framework of quantitative metrics for evaluating data composition quality in LLM pre-training, providing a systematic approach to measure and optimize training data composition.
    
    \item \textbf{Multi-Bucket Method}: We introduce a novel multi-bucket data composition strategy that moves beyond the fixed-length paradigm, demonstrating superior trade-off performance across  proposed quality metrics while maintaining computational efficiency.
    
    \item \textbf{Empirical Validation}: Through extensive experiments, we demonstrate that our method not only optimizes data composition efficiency, but also leads to improved the performance of LLM across various standard benchmarks.
\end{itemize}
\section{Related Work}

In traditional pre-training, individual documents were treated as independent training instances, padded with [PAD] tokens to enable batch processing. 
While effective for smaller models, this approach becomes impractical for modern LLMs with terabyte-scale datasets due to excessive padding overhead.

Initially, most large-scale models~\cite{zhang2022opt,touvron2023llama,le2023bloom,dubey2024llama,yang2024qwen2}, adopted a straightforward fixed-length approach: concatenating documents and segmenting them into sequences matching the model's context window. 
While computationally efficient, this approach often compromises document integrity and semantic coherence, potentially limiting the model's ability to learn long-range dependencies.

Recent works have attempted to address these limitations, albeit with their own constraints. 
Ding et al.~\cite{ding2024fewer} proposed a Best-Fit-Decreasing algorithm that reduces unnecessary truncation, but inherits other limitations of fixed-length approaches such as forced document concatenation and the inability to handle varying sequence lengths efficiently. 
Hadi et al.~\cite{pouransari2024dataset} introduced a more flexible length-based bucketing approach with binomial decomposition. 
While this method avoids fixed-length constraints and eliminates padding and concatenation issues, it introduces extensive document fragmentation, as most documents undergo truncation through the decomposition process.
\section{Preliminary}
\label{preliminary}

To systematically evaluate and compare different data composition strategies, we propose a quantitative framework for assessing training data quality. Our framework is guided by three key objectives:
\begin{itemize}
    \item \textbf{Information Preservation}: Minimizing document truncation to retain the original semantic content and preserve long-range dependencies in the training data.
    
    \item \textbf{Semantic Coherence}: Limiting document concatenation within individual training samples to reduce the noise introduced by artificial boundaries during auto-regressive training.
    
    \item \textbf{Computational Efficiency}: Optimizing the use of padding tokens to maximize computational resources and training throughput.
\end{itemize}
Below, we formally define these objectives as quantitative metrics and analyze their implications for LLM pre-training performance.

We introduce in detail the calculation of the proposed metrics, including: padding ratio, truncation ratio, and concatenate ratio.

(1) Padding Ratio: indicates the proportion of the padding token in the processed training data, which is defined as:
\begin{equation}
    r_{pad} = \frac{{\sum_{i=1}^{N} count(i)}} {\sum_{i=1}^{N} len(i)} \ ,
\end{equation}
where $N$ represents the total number of input data, $count(i)$ represents the number of [PAD] in the $i$-th data, and $len(i)$ refers to the length of the $i$-th data.
The lower the $r_{pad}$, the more effective training.

(2) Truncation Ratio: the ratio of the original training data to be split, which is defined as:
\begin{equation}
    r_{tru} = \frac{{\sum_{i=1}^{M} J(i)}} {M} \ ,
\end{equation}
where $M$ represents the total number of original documents, and $J(i)$ represents that if $i$-th original documents is split, $J(i)$ is 1. 
The lower the $r_{tru}$, the more data is fully trained, and less information is lost.

(3) Concatenate Ratio: the ratio of original data contained in a piece of processed training data. 
\begin{equation}
    r_{cat} = \frac{M} {C} \ ,
\end{equation}
where $M$ represents the total number of original documents, and $C$ represents the number of the training samples.
The lower the $r_{cat}$, the less the number of original data concatenated in the training data and the less noise during training.
\section{Bucketing Algorithm Design}

Optimizing all three metrics ($r_{pad}$, $r_{trunc}$, and $r_{cat}$) simultaneously presents inherent challenges, as these metrics often compete with each other. 
A longer fixed sequence length typically increases concatenation ratio while reducing truncation, and vice versa. 
To address this trade-off, we explore a multi-bucket approach for organizing pre-training data.

\paragraph{Naive Bucketing Method} We first consider a straightforward approach that organizes documents into pre-defined bucket sizes (2048, 4096, 8192, 16384 tokens):
\begin{itemize}
   \item Documents $\leq$ 2048 tokens go to the 2048-token fixed bucket.
   \item Documents of 2049-4096 tokens go to the 4096-token fixed bucket.
   \item Similarly for 4097-8192 and 8193-16384 token ranges.
   \item Documents in each bucket are concatenated and split into fixed-length segments.
\end{itemize}

However, this naive approach suffers from a critical flaw: excessive truncation. 
Most documents, except those perfectly matching bucket sizes or ideal concatenation scenarios, lose information during the fixed-length segmentation process. 
This substantial information loss motivates the need for a more sophisticated data organization strategy that better preserves document integrity while maintaining training efficiency.

\begin{algorithm}[t]
\caption{BucketLLM}
\label{alg:bucketing_final_optimized_1}
\begin{algorithmic}[1]
\REQUIRE All Documents $\mathcal{DS}$, preset bucket set $B$ with capacities, padding threshold $P$, Pool Size $S$
\ENSURE All Training buckets $\mathcal{TB}$
\STATE initialize a empty document pool $D$
\FOR{each $d_i$ in $\mathcal{DS}$}
    \STATE Add $d_i$ to $D$
    \IF{ $len(D) >= S$}
        \STATE $T, D \gets Generate\_a\_Bucket(D, B, P)$
    \STATE add $T$ to $\mathcal{TB}$
    \ENDIF
\ENDFOR
\RETURN $\mathcal{TB}$
\end{algorithmic}
\end{algorithm}

\paragraph{BucketLLM-v0: Improved Bucketing Algorithm}
To address the limitations of the naive approach, we propose an improved bucketing method that adaptively organizes documents using a greedy strategy. 
The key innovation is allowing flexible bucket allocation: instead of strict length-based assignments, documents can be placed in larger buckets when beneficial for minimizing truncation and maximizing space utilization. 
The algorithm processes documents in batches as a document pool from a globally shuffled dataset:

\begin{enumerate}
   \item Initial Organization: Documents in the current pool are sorted in descending order by length to prioritize placing larger documents.
   
   \item Bucket Assignment: For each document, we select the smallest viable bucket that can accommodate it completely.
   
   \item Space Optimization: 
   \begin{itemize}
       \item Long documents exceeding the maximum bucket size are split, with the remainder returning to the pool.
       \item When a bucket cannot fit more complete documents, if remaining space exceeds a padding threshold, it is filled with a chunk from the shortest available document.
       \item Small remaining spaces below the threshold are filled with padding tokens.
   \end{itemize}
\end{enumerate}

%This adaptive approach significantly reduces truncation compared to the naive method while maintaining efficient bucket utilization.

% This approach significantly reduces the truncation ratio ($r_{trunc}$) compared to the naive method, as it allows for more flexible document placement. 
% The concatenation ratio ($r_{cat}$) is generally kept low. 
% The padding ratio ($r_{pad}$) is controlled by the padding threshold, allowing for a balance between space utilization and computational efficiency.
% By employing this greedy strategy, we can better optimize all three metrics simultaneously compared to the naive approach, leading to more efficient data composition for large language model pre-training, and we name this method as BucketLLM-v0.

% However, this greedy strategy has some characteristics and limitations worth noting.
% The current greedy approach selects the longest remaining document that fits into the current bucket. This can lead to situations where, after placing the longest fitting document, a significant amount of space remains unused, but no other document can fit entirely. 
% This potentially increases the padding ratio and truncation ratio, which may not be ideal.

This method demonstrates several improvements over the naive approach. 
It significantly reduces the truncation ratio ($r_{trunc}$) through flexible document placement, while maintaining a low concatenation ratio ($r_{cat}$). The padding ratio ($r_{pad}$) is effectively managed through the padding threshold parameter, enabling a tunable balance between space utilization and computational efficiency.

However, BucketLLM-v0's greedy strategy reveals certain limitations. By prioritizing the longest fitting document for each bucket, it can create suboptimal space allocation scenarios. Specifically, after placing a long document, the remaining space may be too small for any complete document, yet too large to justify padding. 
Moreover, this strategy tends to favor larger buckets, as they offer more flexibility in document placement, potentially leading to an imbalanced distribution of training sequences across different length buckets. 
This bias towards longer sequences not only affects training efficiency but may also impact the model's ability to handle inputs of varying lengths effectively. These observations motivate the need for a more sophisticated document selection strategy that can better balance bucket utilization across different sequence lengths.

\begin{algorithm}[t]
\caption{Generate\_a\_Bucket}
\label{alg:bucketing_final_optimized_2}
\begin{algorithmic}[1]
\REQUIRE A Documents Pool $D$, preset bucket set $B$ with capacities, padding threshold $P$
\ENSURE A Training bucket $T$, updated Documents Pool D
\STATE Sort $D$ in descending order by length 
%\WHILE{$D$ is not empty} 
    \STATE $s \gets$ select bucket size from $B$ to fit $D_1$, initialized the empty bucket  T
    \FOR{each $d_i$ in $D$}
        \IF{$len(d_i) \leq s$}
            %\STATE $\tau \gets d_i$
            \STATE $r \gets s - len(d_i)$
            \IF{$\min_{d \in D} len(d) > r > P \cdot \text{bucket capacity} $}
                \STATE Find $a, b \in D$ where $len(a) + len(b) \leq r$ and $len(a) + len(b) > len(d_i)$
                \IF{$a, b$ found}
                    \STATE Add $a, b$ to bucket T
                    \STATE $s \gets s - (len(a) + len(b))$
                    \STATE $a, b \gets \emptyset$
                \ELSE
                    \STATE Add $d_i$ to bucket T
                    \STATE $s \gets r$
                \ENDIF
            \ELSE
                \STATE Add $d_i$ to bucket T
                \STATE $s \gets r$
            \ENDIF
            \STATE $\tau \gets \emptyset$
        \ELSIF{bucket is empty} 
            \STATE Add $d_i[0:b]$ to bucket T
            \STATE $d_i \gets d_i[b:]$  \COMMENT{Update document}
        \ENDIF   
    \ENDFOR
    \STATE Shuffle the bucket T
    \IF{$r / \text{bucket capacity} > P$}
        \STATE Add $D[-1][0:r]$ to bucket T \COMMENT{Chunk from shortest document}
        \STATE $D[-1] \gets D[-1][r:]$  
    \ELSE
        \STATE Add $[pad\_id] \times r$ to bucket T
    \ENDIF
% \STATE Remove empty documents from $D$
\RETURN $T$, $D$
%\ENDWHILE
\end{algorithmic}
\end{algorithm}

\paragraph{BucketLLM: Final Optimized Bucketing Algorithm}

Building upon the greedy approach described earlier, we propose a further optimized algorithm that incorporates \textbf{heuristic strategies} to address the limitations of the previous method. 
This final algorithm aims to minimize padding and  truncation while maximizing the utilization of bucket space. 
The key improvement lies in its handling of scenarios where the remaining space is too small for the next largest document but too large to be efficiently padded.

Compared with the workflow in the above BucketLLM-v0, this algorithm has a more detailed solution in \textit{Space Optimization} step: when a document $\tau$ is added and the remaining space $r$ is smaller than the shortest document but larger than the padding threshold, apply the following heuristic:
    \begin{itemize}
        \item Search for two shorter documents $a$ and $b$ such that $len(a) + len(b) \leq r$ and $len(a) + len(b) > len(\tau)$.
        \item If such a pair is found, add $a$ and $b$ to the bucket instead of $\tau$.
        \item If no such pair is found, proceed with adding $\tau$.
    \end{itemize}

This heuristic approach provides several advantages:
(1) Reduced Padding: By finding pairs of shorter documents to fill larger gaps, we significantly reduce the amount of padding required, thus decreasing $r_{pad}$.
(2) Improved Space Utilization: The algorithm makes more efficient use of the available bucket space, potentially increasing the amount of useful data in each training sample.

Mathematically, we can express the condition for applying the heuristic as:

\[
\min_{d \in D} len(d) > r > P \cdot \text{bucket\_capacity} \ ,
\]
where $D$ is the set of remaining documents, $r$ is the remaining space in the current bucket, and $P$ is the padding threshold.

The optimization goal for selecting documents $a$ and $b$ can be expressed as:

\begin{equation}
\begin{split}
     &\max_{a,b \in D} (len(a) + len(b)) \text{ subject to } len(a) + len(b) \leq r \ , \\ 
    &\text{ and } len(a) + len(b) > len(\tau) 
\end{split}
\end{equation}

The detailed pseudocode in Algorithm \ref{alg:bucketing_final_optimized_1} and Algorithm \ref{alg:bucketing_final_optimized_2} illustrates the step-by-step process of our method to optimized bucket creation.

Our algorithm is designed to work with a preset bucket set $B$ (e.g., 1024, 2048, 4096, ...), which allows for flexibility in defining bucket capacities based on specific model or hardware requirements. 
The padding threshold $P$ provides a tunable parameter to balance between space utilization and computational efficiency.

Overall, the key features and improvements of our method are as follows:
\begin{itemize}

    \item Document Pool Management: We introduce a document pool of size $S$, allowing for more flexible and efficient handling of very large datasets. This approach ensures a consistent flow of data processing and memory efficiency.
    
    \item Adaptive Bucket Size Selection: The algorithm dynamically selects the appropriate bucket size based on the current longest document in the pool, ensuring efficient use of different bucket capacities.
    
    \item Optimized Bucket Filling: Our method implements an improved greedy strategy with heuristic optimization. It aims to maximize bucket utilization while minimizing padding and truncation.
\end{itemize}

This method effectively balances the three key metrics we introduced earlier, and could achieve a more efficient and flexible data composition process for LLM pre-training. 
This approach not only improves the quality of training data by preserving document integrity where possible but also enhances computational efficiency by optimizing bucket utilization and minimizing unnecessary padding and concatenation.

\begin{table*}
\centering
\resizebox{0.95\textwidth}{!}{
\begin{tabular}{c|c|c|c|c|c|c|c|c}
\toprule[1.5pt]
  & Fixed-1024 & Fixed-2048 & Fixed-4096 & Fixed-8192 & Fixed-16384 & DD &  \textbf{BucketLLM-v0} & \textbf{BucketLLM }\\
\midrule
 $ r_{pad} \downarrow $  & 0.01\% &0.01\%  & 0.02\% & 0.04\% & 0.08\% & 0.0\% & 0.37\% & 0.12\% \\
\midrule
$r_{tru} \downarrow$ & 61.26\% & 38.39\% & 21.45\% & 11.51\% & 5.96\% & 99.9\% &0.19\%  & 0.28\% \\
\midrule
$r_{cat} \downarrow$ & 1.03 & 2.05 & 4.11 & 8.20 & 16.40  & 1.0 & 3.78 & 2.81 \\
\bottomrule[1.5pt]
\end{tabular}
}
\caption{Comparison of Data Composition Quality Metrics across Different Methods. $r_{pad}$: padding ratio, $r_{tru}$: truncation ratio, $r_{cat}$: concatenation ratio (lower is better for all metrics).}
\label{tab:data-performance}
\end{table*}

\section{Experiments}

To comprehensively evaluate our proposed multi-bucket data composition method, we conduct experiments to address the following research questions:

\textbf{RQ1: How does our multi-bucket method improve data composition quality?} 
To answer this question, we evaluate our method against conventional fixed length and SOTA data composition strategies using the proposed metrics: padding ratio, truncation ratio and concatenation ratio.

\textbf{RQ2: Does our proposed method lead to better model performance in LLM pre-training?} 
To validate the effectiveness of our approach, we pre-train language models using our multi-bucket data composition method and evaluate them on widely used benchmarks. 
Meanwhile, we particularly focus on examining the model's capability in capturing long-range dependencies and processing longer sequences.
%Furthermore, we demonstrate that our method effectively supports both data scaling and model size scaling, enabling efficient pre-training across different model capacities and training corpus sizes.

\subsection{Experimental Setup}

\paragraph{Datasets}
We utilize FineWebEdu~\cite{penedo2024fineweb}, a high-quality pre-training dataset derived from CommonCrawl. 
We conduct our experiments on a representative subset consisting of 100B tokens (approximately 98M documents), and this subset maintains the original dataset's quality characteristics and diversity, making it suitable for evaluating the data composition method.

\paragraph{Baselines}
We compare our method with the following baselines.
\begin{itemize}
    \item \textbf{Fixed Length}: The standard approach of concatenating all documents and then splitting them into fixed-length samples for training such as Fixed-1024, Fixed-2048, Fixed-4096, Fixed-8192, and Fixed-16384. 
    
    \item \textbf{DD}~\cite{pouransari2024dataset}: A state-of-the-art data composition method that groups sequences by length into document-specific buckets and employs curriculum-based sampling during training. 
\end{itemize}
In our method, we pre-define five buckets with commonly used sequence lengths: 1024, 2048, 4096, 8192, and 16384.

\subsection{RQ1: Data Composition Quality}
To evaluate the effectiveness of our proposed multi-bucket data composition strategy, we conduct a comprehensive analysis of data organization quality across different methods. 
Specifically, we compare our approach against five fixed-length baselines (Fixed-1024, Fixed-2048, Fixed-4096, Fixed-8192, and Fixed-16384) and the DD method measuring their performance in terms of three proposed key metrics: padding ratio, truncation ratio, and concatenation ratio. 
%These metrics provide quantitative insights into how effectively each method handles the inherent variability in document lengths while maintaining computational efficiency.

Table~\ref{tab:data-performance} presents the comparison of different methods across three data composition quality metrics. 
For fixed-length methods, we observe clear trade-offs: shorter sequence lengths (e.g., Fixed-1024) lead to lower padding ratios but higher truncation ratios, while longer sequences (e.g., Fixed-16384) reduce truncation at the cost of increased concatenation meanwhile significantly slower training speed due to long context attention complexity. 
The DD method achieves minimal padding and concatenation but at the expense of extremely high truncation (99.9\%), which could potentially harm model performance by breaking the natural document boundaries and coherent semantic units.
Our proposed BucketLLM accepts a slightly higher padding ratio (0.12\%) compared to fixed-length methods and DD, but this small overhead brings substantial benefits: it dramatically reduces the truncation ratio to 0.28\% while maintaining a relatively low concatenation ratio (2.81). 
This represents a well-balanced trade-off, as the minimal increase in padding has negligible computational cost, while the reduced truncation better preserves document integrity and the controlled concatenation better preserve the natural document boundaries and semantic coherence. 

\begin{figure}
\centering
\includegraphics[width=0.44\textwidth]{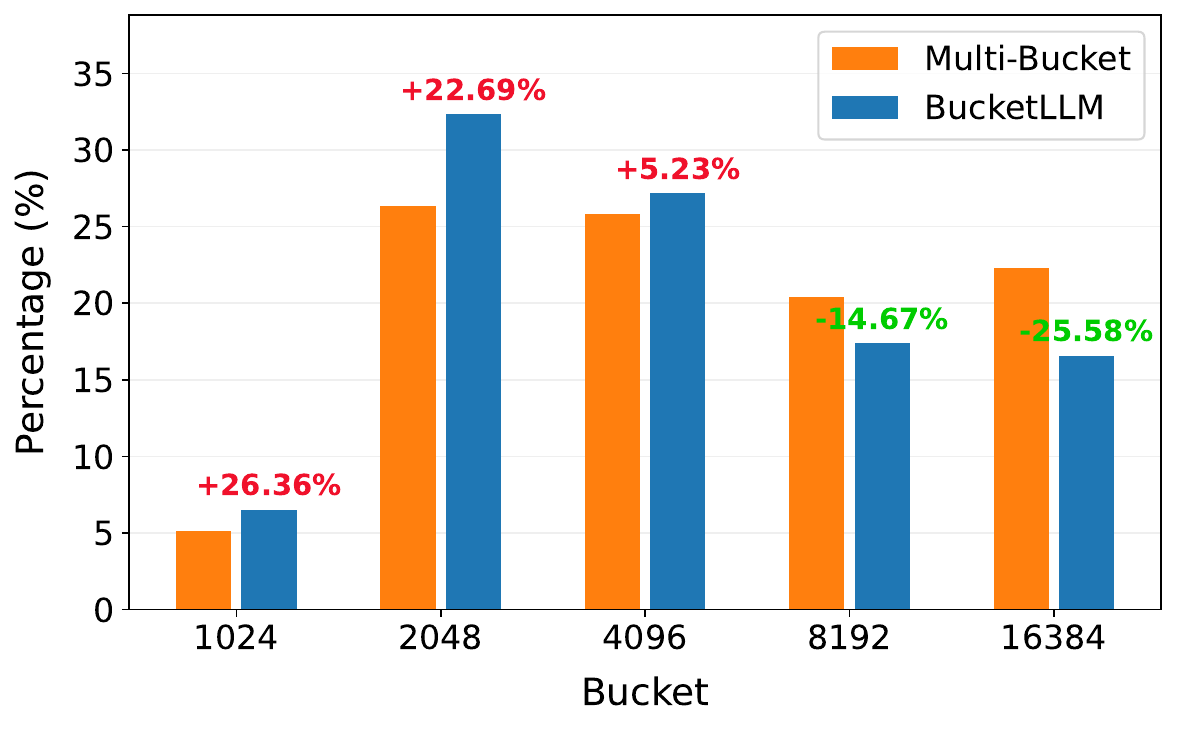}
\caption{Different Bucket Distribution of BucketLLM-v0 and BucketLLM.}
\label{fig:distribution} 
\end{figure}

In addition, we compare our preliminary version BucketLLM-v0, which could  help demonstrate the effectiveness and insights of the evolution of our approach.
Figure~\ref{fig:distribution} shows the sequence length distribution between our two methods. 
BucketLLM demonstrates a more balanced distribution with a notable shift towards shorter sequences compared to BucketLLM-v0.
Specifically, BucketLLM increases the proportion of sequences in shorter buckets (1024-4096 tokens) while reducing the percentage in longer buckets (8192-16384 tokens). 
Combined with the metrics in Table~\ref{tab:data-performance}, this distribution shift demonstrates that BucketLLM achieves a better balance between quality and efficiency: while both methods achieve low padding and truncation ratios, BucketLLM's better utilization of shorter sequences leads to a reduced concatenation ratio (2.81 vs 3.78). 
Importantly, its preference for shorter sequences leads to substantially improved training efficiency.

\begin{table*}
\centering
\resizebox{0.8\textwidth}{!}{
\begin{tabular}{c|c|c|c|c|c}
\toprule[1.5pt]
 & DD & Fixed-2048 & Fixed-4096 & Fixed-8192 & BucketLLM \\
 
\midrule
Arc-challenge  & 0.2920 & \underline{0.3000} & 0.2920 & 0.2760 & \textbf{0.3140} (+4.67\%)  \\
\midrule
Arc-easy & 0.6570 & \underline{0.6590} & 0.6480 & 0.6420  & \textbf{0.6640} (+0.76\%)  \\
\midrule
Commonsense-qa & \underline{0.3790} & 0.3610 & 0.3530 & 0.3730  & \textbf{0.3870} (+2.11\%)  \\
\midrule
Hellaswag & 0.3860 & \underline{0.4030} & 0.3960 & 0.3810  & \textbf{0.4060} (+0.74\%)  \\
\midrule
Mmlu-average & \underline{0.3118} & 0.3098 & 0.3077 & 0.2993 & \textbf{0.3165} (+1.51\%)   \\
\midrule
Openbookqa & 0.2420 & 0.2420 & \underline{0.2580} & 0.2480  & \textbf{0.2620} (+1.55\%)  \\
\midrule
Piqa & 0.6880 & \underline{0.6970} & 0.6880 & 0.6680 & \textbf{0.7020} (+0.72\%)  \\
\midrule
\midrule
Average & 0.4223 & \underline{0.4245} & 0.4204 & 0.4125 & \textbf{0.4359} (+2.69\%)  \\
\bottomrule[1.5pt]
\end{tabular}
}
\caption{ Performance on   Benchmarks. Numbers show accuracy scores for each task. Best results are in \textbf{bold}, and the best baseline results are \underline{underlined}. The percentage improvements of BucketLLM are relative to the best baseline performance.}
\label{tab:evalperformance}
\end{table*}

\subsection{RQ2: Pre-training LLM Performance}

\paragraph{Experimental Settings}
To evaluate the effectiveness of different data composition methods on model performance, we conduct extensive pre-training experiments using a 1B parameter model based on the Llama3.1 architecture~\cite{dubey2024llama}. 
For comparison, we select three widely used fixed-length baselines (Fixed-2048, Fixed-4096, and Fixed-8192) and the DD method. 
All models are trained using the DeepSpeed framework on 8 nodes, each equipped with 8 NVIDIA A800 GPUs (64 GPUs in total). 
We maintain consistent training configurations across all methods, including an initial learning rate of 2e-4.
Notably, we implement a bucket sampling algorithm where buckets are sampled proportionally to their token counts at each training step, and the sampling algorithm could ensure that data across all GPUs of the same step consistently reside within the same bucket, facilitating efficient training.

\paragraph{Training Efficiency}
Table~\ref{tab:spped} presents the distribution of training data across different bucket sizes and their corresponding training throughput. 
Our multi-bucket approach shows two key advantages in terms of training efficiency. First, we achieve a favorable data distribution where 55.44\% of the training data falls into buckets of 4,096 tokens or shorter. 
And these shorter sequences demonstrate substantially higher training throughput: compared to the 8,192 sequence length (which has become a common choice in recent LLM training).
While longer sequences (16,384) show reduced throughput (-18.74\%), they only account for 17.39\% of the training data. 
This distribution-aware training strategy effectively balances the trade-off between sequence length and computational efficiency, leading to improved overall training throughput compared to fixed-length approaches.

\begin{table}
\centering
\resizebox{0.46\textwidth}{!}{
\begin{tabular}{c|c|c|c}
\toprule[1.5pt]
bucket &data ratio & batch size& speed (tokens/s) \\
\midrule
1,024 & 6.52\% & 48 & 16,793 (\red{+22.62\%}) \\
2,048 & 16.58\% & 24 & 16,263 (\red{+18.74\%})  \\
4,096 & 32.34\% & 12 & 15,505  (\red{+13.21\%}) \\
8,192 & 27.17\% & 6& 13,696 (+0.00\%)\\
16,384 & 17.39\% & 3 & 11,129 (\blue{-18.74\%}) \\
\bottomrule[1.5pt]
\end{tabular}
}
\caption{Training setting across Different Bucket Sizes. Speed improvements are relative to the widely used 8,192 context length.}
\label{tab:spped}
\end{table}

\paragraph{Benchmark Evaluation}

We evaluate each model on a comprehensive set of standard benchmarks under Lighteval framework\footnote{https://github.com/huggingface/lighteval}, including arc-challenge~\cite{clark2018think}, arc-easy~\cite{clark2018think}, commonsense-qa~\cite{talmor2019commonsenseqa}, hellaswag~\cite{zellers2019hellaswag}, mmlu-average~\cite{hendrycksmeasuring}, openbook-qa~\cite{mihaylov2018can}, and piqa~\cite{bisk2020piqa}.
We report accuracies for each category.

Table~\ref{tab:evalperformance} presents the evaluation results across seven standard NLP benchmarks. 
Our BucketLLM consistently outperforms all baseline methods across all tasks, achieving an average improvement of 2.69\% over the best baseline (Fixed-2048). 
Notably, we observe substantial improvements on more challenging tasks, such as Arc-challenge (+4.67\%) and Commonsense-qa (+2.11\%). 
Since these benchmarks primarily involve short-text understanding and reasoning, it is unsurprising that Fixed-2048 emerges as the strongest baseline among fixed-length approaches, as it better matches the natural length distribution of these tasks. 
In contrast, Fixed-8192 shows the worst performance, likely due to excessive concatenation introducing cross-document noise during training. 
Our multi-bucket approach effectively addresses this challenge by adaptively handling sequences of different lengths, leading to more robust performance across the full benchmark suite while maintaining training efficiency.
\begin{figure}
\centering
\includegraphics[width=0.41\textwidth]{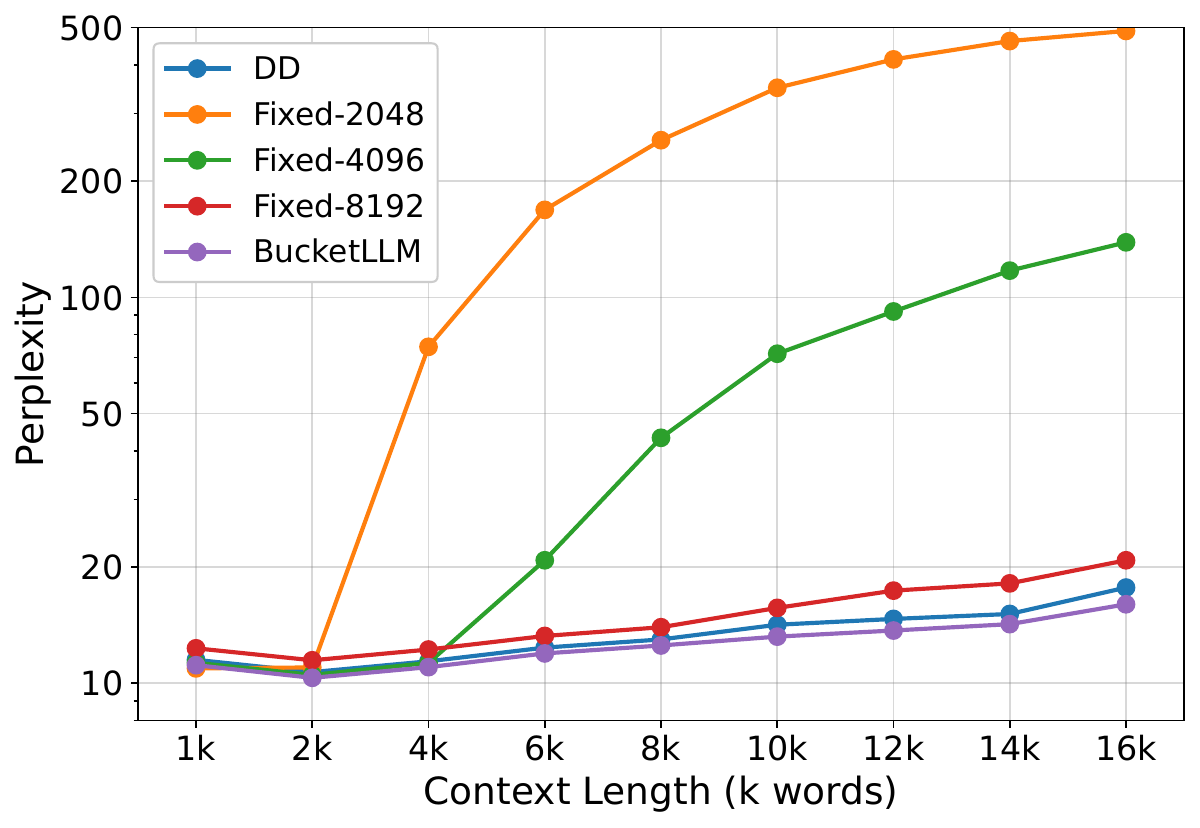}
\caption{PPL Performance with Different Context Length.}
\label{fig:ppl} 
\end{figure}

\paragraph{Long Context Modeling Evaluation}
We evaluate models' capability in handling texts of varying lengths by measuring perplexity across different context windows, from 1k to 16k, as shown in Figure~\ref{fig:ppl}. 
Our BucketLLM and the other bucket method DD demonstrates superior performance across the entire length spectrum, maintaining the lowest perplexity for both short and long contexts. 
This consistent performance can be attributed to the adaptive bucket strategy, which exposes the model to an appropriate mix of sequence lengths during training. 
Notably, while Fixed-2048 showed strong performance in previous short-text benchmarks, it struggles significantly with longer contexts, showing rapidly degrading perplexity beyond its training length. 

The experimental results demonstrate that our adaptive multi-bucket strategy effectively balances the trade-offs in LLM training: by intelligently distributing sequences across different length buckets, BucketLLM not only accelerates training through efficient processing of shorter sequences, but also maintains robust modeling capabilities across both short and long contexts, leading to superior performance in both standard benchmarks and long-text understanding tasks.

\begin{figure}
\centering
\includegraphics[width=0.42\textwidth]{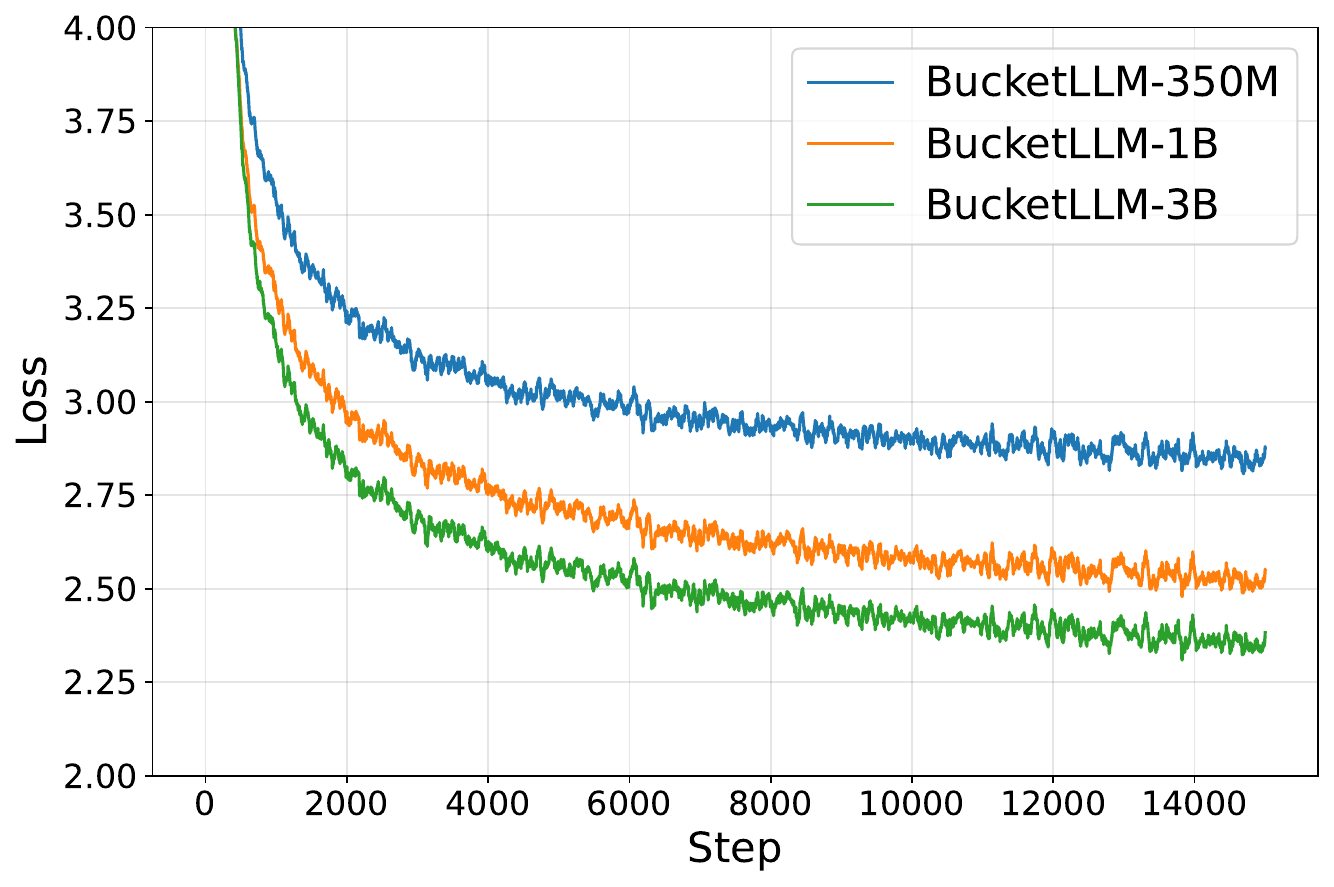}
\caption{Loss Curve with Different Model Parameters.}
\label{fig:loss} 
\end{figure}

\begin{figure}
\centering
\includegraphics[width=0.42\textwidth]{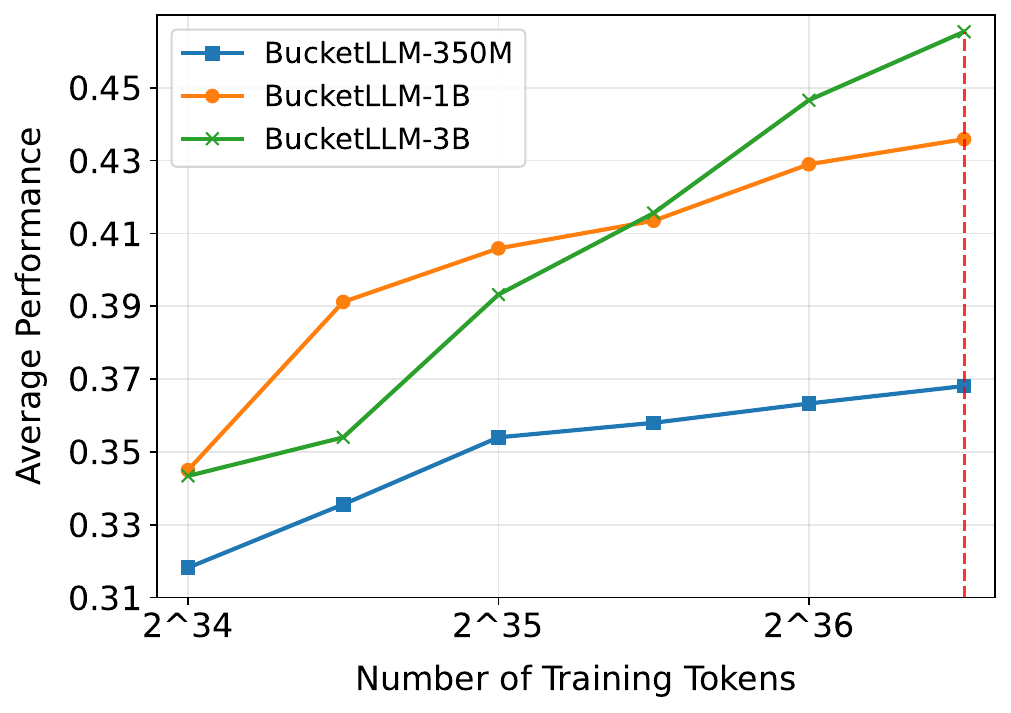}
\caption{Data Scaling and Model Scaling (red line).}
\label{fig:datascaling} 
\end{figure}

\subsection{Further Scaling Analysis}

Figures~\ref{fig:loss} and~\ref{fig:datascaling} present comprehensive analyses of our method's scaling behavior across different model sizes (350M, 1B, and 3B parameters) and training data scales. 
First, the loss curves (Figure~\ref{fig:loss}) demonstrate consistent and stable training dynamics across all model sizes, with the larger models (3B) achieving lower loss values as expected.
The convergence tendency suggests that our multi-bucket approach effectively supports stable training regardless of model scale.

More importantly, Figure~\ref{fig:datascaling} reveals the interaction between model size and data scaling. 
When increasing the number of training tokens from $2^{34}$ to $2^{36}$, we observe distinct scaling patterns: the 350M model shows limited improvement with more data, suggesting it reaches its capacity limit earlier. 
In contrast, both 1B and 3B models demonstrate stronger scaling behavior, with the 3B model showing the most substantial performance gains as training data increases. 
This indicates that our bucket-based training strategy effectively supports both data and model scaling, allowing larger models to better utilize increased training data for improved performance.

These findings suggest that BucketLLM's adaptive sequence length strategy is particularly beneficial for scaling up model training, as it maintains efficient training while enabling models to effectively leverage both increased model capacity and training data.

\section{Conclusion}
In this paper, we study the data composition problem in LLM pre-training by introducing a systematic approach to evaluate and optimize training data composition. 
Our key contributions include establishing three quantitative metrics - padding ratio, truncation ratio, and concatenation ratio - providing a comprehensive framework for measuring data composition quality. 
We further propose BucketLLM, an adaptive multi-bucket data composition method that transcends the traditional fixed-length paradigm, which demonstrates superior performance in both data composition quality and model training.  
Our method shows substantial improvements in training efficiency and model performance across different scales and benchmarks. Our work provides valuable insights into efficient LLM training through better data organization strategies.

% In this paper, we first design three quantitative indicators, namely padding ratio, truncation ratio, and concatenate ratio, to measure the quality of data composition.
% We  propose a novel multi-bucket data composition method that moves beyond the fixed-length paradigm.
% Compared with existing  fixed length training methods, our method can not only obtain more high quality pertaining data samples in our evaluation metrics but also achieve efficient and effective training performance for LLMs. 
%take into account the training effect of short texts, the modeling ability of long contexts, and the overall training speed.

\clearpage

\bibliographystyle{named}
\bibliography{ijcai25}

\begin{thebibliography}{}

\bibitem[\protect\citeauthoryear{Bisk \bgroup \em et al.\egroup }{2020}]{bisk2020piqa}
Yonatan Bisk, Rowan Zellers, Jianfeng Gao, Yejin Choi, et~al.
\newblock Piqa: Reasoning about physical commonsense in natural language.
\newblock In {\em Proceedings of the AAAI conference on artificial intelligence}, volume~34, pages 7432--7439, 2020.

\bibitem[\protect\citeauthoryear{Brown \bgroup \em et al.\egroup }{2020}]{brown2020language}
Tom Brown, Benjamin Mann, Nick Ryder, Melanie Subbiah, Jared~D Kaplan, Prafulla Dhariwal, Arvind Neelakantan, Pranav Shyam, Girish Sastry, Amanda Askell, et~al.
\newblock Language models are few-shot learners.
\newblock {\em Advances in neural information processing systems}, 33:1877--1901, 2020.

\bibitem[\protect\citeauthoryear{Clark \bgroup \em et al.\egroup }{2018}]{clark2018think}
Peter Clark, Isaac Cowhey, Oren Etzioni, Tushar Khot, Ashish Sabharwal, Carissa Schoenick, and Oyvind Tafjord.
\newblock Think you have solved question answering? try arc, the ai2 reasoning challenge.
\newblock {\em arXiv preprint arXiv:1803.05457}, 2018.

\bibitem[\protect\citeauthoryear{DeepSeek-AI}{2024}]{deepseekai2024deepseekv3technicalreport}
DeepSeek-AI.
\newblock Deepseek-v3 technical report.
\newblock {\em arXiv preprint arXiv:2412.19437}, 2024.

\bibitem[\protect\citeauthoryear{Ding \bgroup \em et al.\egroup }{2024}]{ding2024fewer}
Hantian Ding, Zijian Wang, Giovanni Paolini, Varun Kumar, Anoop Deoras, Dan Roth, and Stefano Soatto.
\newblock Fewer truncations improve language modeling.
\newblock {\em arXiv preprint arXiv:2404.10830}, 2024.

\bibitem[\protect\citeauthoryear{Dubey \bgroup \em et al.\egroup }{2024}]{dubey2024llama}
Abhimanyu Dubey, Abhinav Jauhri, Abhinav Pandey, Abhishek Kadian, Ahmad Al-Dahle, Aiesha Letman, Akhil Mathur, Alan Schelten, Amy Yang, Angela Fan, et~al.
\newblock The llama 3 herd of models.
\newblock {\em arXiv preprint arXiv:2407.21783}, 2024.

\bibitem[\protect\citeauthoryear{Hendrycks \bgroup \em et al.\egroup }{2020}]{hendrycksmeasuring}
Dan Hendrycks, Collin Burns, Steven Basart, Andy Zou, Mantas Mazeika, Dawn Song, and Jacob Steinhardt.
\newblock Measuring massive multitask language understanding.
\newblock In {\em International Conference on Learning Representations}, 2020.

\bibitem[\protect\citeauthoryear{Le~Scao \bgroup \em et al.\egroup }{2023}]{le2023bloom}
Teven Le~Scao, Angela Fan, Christopher Akiki, Ellie Pavlick, Suzana Ili{\'c}, Daniel Hesslow, Roman Castagn{\'e}, Alexandra~Sasha Luccioni, Fran{\c{c}}ois Yvon, Matthias Gall{\'e}, et~al.
\newblock Bloom: A 176b-parameter open-access multilingual language model.
\newblock 2023.

\bibitem[\protect\citeauthoryear{Mihaylov \bgroup \em et al.\egroup }{2018}]{mihaylov2018can}
Todor Mihaylov, Peter Clark, Tushar Khot, and Ashish Sabharwal.
\newblock Can a suit of armor conduct electricity? a new dataset for open book question answering.
\newblock In {\em Proceedings of the 2018 Conference on Empirical Methods in Natural Language Processing}, pages 2381--2391, 2018.

\bibitem[\protect\citeauthoryear{Ouyang \bgroup \em et al.\egroup }{2022}]{ouyang2022training}
Long Ouyang, Jeffrey Wu, Xu~Jiang, Diogo Almeida, Carroll Wainwright, Pamela Mishkin, Chong Zhang, Sandhini Agarwal, Katarina Slama, Alex Ray, et~al.
\newblock Training language models to follow instructions with human feedback.
\newblock {\em Advances in neural information processing systems}, 35:27730--27744, 2022.

\bibitem[\protect\citeauthoryear{Penedo \bgroup \em et al.\egroup }{2024}]{penedo2024fineweb}
Guilherme Penedo, Hynek Kydl{\'\i}{\v{c}}ek, Anton Lozhkov, Margaret Mitchell, Colin Raffel, Leandro Von~Werra, Thomas Wolf, et~al.
\newblock The fineweb datasets: Decanting the web for the finest text data at scale.
\newblock {\em arXiv preprint arXiv:2406.17557}, 2024.

\bibitem[\protect\citeauthoryear{Pouransari \bgroup \em et al.\egroup }{2024}]{pouransari2024dataset}
Hadi Pouransari, Chun-Liang Li, Jen-Hao~Rick Chang, Pavan Kumar~Anasosalu Vasu, Cem Koc, Vaishaal Shankar, and Oncel Tuzel.
\newblock Dataset decomposition: Faster llm training with variable sequence length curriculum.
\newblock In {\em Proceedings of the Annual Conference on Neural Information Processing Systems}, 2024.

\bibitem[\protect\citeauthoryear{Rae \bgroup \em et al.\egroup }{2021}]{rae2021scaling}
Jack~W Rae, Sebastian Borgeaud, Trevor Cai, Katie Millican, Jordan Hoffmann, Francis Song, John Aslanides, Sarah Henderson, Roman Ring, Susannah Young, et~al.
\newblock Scaling language models: Methods, analysis \& insights from training gopher.
\newblock {\em arXiv preprint arXiv:2112.11446}, 2021.

\bibitem[\protect\citeauthoryear{Talmor \bgroup \em et al.\egroup }{2019}]{talmor2019commonsenseqa}
Alon Talmor, Jonathan Herzig, Nicholas Lourie, and Jonathan Berant.
\newblock Commonsenseqa: A question answering challenge targeting commonsense knowledge.
\newblock In {\em Proceedings of the 2019 Conference of the North}, page 4149. Association for Computational Linguistics, 2019.

\bibitem[\protect\citeauthoryear{Touvron \bgroup \em et al.\egroup }{2023}]{touvron2023llama}
Hugo Touvron, Louis Martin, Kevin Stone, Peter Albert, Amjad Almahairi, Yasmine Babaei, Nikolay Bashlykov, Soumya Batra, Prajjwal Bhargava, Shruti Bhosale, et~al.
\newblock Llama 2: Open foundation and fine-tuned chat models.
\newblock {\em arXiv preprint arXiv:2307.09288}, 2023.

\bibitem[\protect\citeauthoryear{Yang \bgroup \em et al.\egroup }{2024}]{yang2024qwen2}
An~Yang, Baosong Yang, Beichen Zhang, Binyuan Hui, Bo~Zheng, Bowen Yu, Chengyuan Li, Dayiheng Liu, Fei Huang, Haoran Wei, et~al.
\newblock Qwen2. 5 technical report.
\newblock {\em arXiv preprint arXiv:2412.15115}, 2024.

\bibitem[\protect\citeauthoryear{Zellers \bgroup \em et al.\egroup }{2019}]{zellers2019hellaswag}
Rowan Zellers, Ari Holtzman, Yonatan Bisk, Ali Farhadi, and Yejin Choi.
\newblock Hellaswag: Can a machine really finish your sentence?
\newblock In {\em Proceedings of the 57th Annual Meeting of the Association for Computational Linguistics}, pages 4791--4800, 2019.

\bibitem[\protect\citeauthoryear{Zhang \bgroup \em et al.\egroup }{2022}]{zhang2022opt}
Susan Zhang, Stephen Roller, Naman Goyal, Mikel Artetxe, Moya Chen, Shuohui Chen, Christopher Dewan, Mona Diab, Xian Li, Xi~Victoria Lin, et~al.
\newblock Opt: Open pre-trained transformer language models.
\newblock {\em arXiv preprint arXiv:2205.01068}, 2022.

\end{thebibliography}

\end{document}